%% file: nv_main.tex
\documentclass[11pt, logo, onecolumn, copyright, colorlinks=true, allcolors=blue]{nvidiatechreport}

\usepackage{graphicx}
\usepackage{subcaption,ragged2e}
\usepackage[round]{natbib}
\usepackage{multirow}
\usepackage{booktabs}
\usepackage{amsmath}
\usepackage{url}
\title{Online Draft Co-Training for Speculative Decoding in Large-Scale, Long-Context RL Post-Training}
\author{Zili Wang, Zhaopeng Qiu, Yuekai Zhang, Shuang Yu, Seonjin Na, Hayate Iso, Junjie Lai\\
{\normalfont\small\texttt{\{ziliw, alexq, yuekaiz, shuangy, sna, hiso, julienl\}@nvidia.com}}
}
\date{2026-9-7}

\begin{document}

\input{sec/abstract}

\maketitle

\input{sec/introduction}

\input{sec/method}
\input{sec/experiments}
\input{sec/related_work}
\input{sec/conclusion}

\bibliography{references}
\bibliographystyle{plainnat}

\input{sec/appendix}

\end{document}

%% file: sec/abstract.tex
\begin{abstract}
\large \textbf{Abstract.} Speculative decoding accelerates rollout generation, which dominates the cost of
reinforcement learning (RL) post-training. Online co-training keeps the draft aligned with the evolving policy, further increasing acceptance length and speedup. However, training large models on long contexts requires context parallelism (CP) and pipeline parallelism (PP), and co-training drafts under this layout poses two obstacles: (1) the draft's branch attention is unsupported by causal CP, and (2) the target features that the draft consumes are spread across PP stages. In this paper, we present the design and implementation of online draft co-training in NeMo-RL. For CP, we extend packed, load-balanced zigzag ring attention by merging
rank-local branch attention with causal main-sequence attention. For PP, TapChannel transports
intermediate target features across stages via a separate path, leaving the pipeline schedule unaffected. Experiments show that co-training keeps the policy's reward and accuracy close to the no-draft baseline while delivering $1.48$--$2.25\times$ rollout speedups and $1.19$--$1.88\times$ end-to-end training speedups across model scales up to 122B.
Our CP design achieves strong scaling at 256K sequence length with significant memory savings over
prior work, and our PP transport incurs modest overhead.
Code can be found \href{https://github.com/NVIDIA-NeMo/RL/issues/3698}{\textbf{here}}.
\end{abstract}

%% file: sec/introduction.tex
\section{Introduction}
\label{sec:introduction}

Reinforcement learning (RL) post-training has become a standard paradigm for
building reasoning and agentic Large Language Models (LLMs)
\citep{shao2024deepseekmath,yu2025dapo,glm5team2026glm5} and its wall-clock time is often dominated by rollout generation~\citep{zhang2026fastgrpo,chen2026respec}. Speculative decoding (SD)~\citep{leviathan2023fast,chen2023accelerating} accelerates generation
without changing the output distribution: a small draft model proposes several tokens,
and the target policy verifies them in parallel.
Recent RL frameworks have therefore integrated SD into their rollout engines,
showing that rollout speedups can translate into end-to-end training acceleration
\citep{liu2025spec,iso2026accelerating, kim2026efficientrollout, shao2026beat, verl_mtp_docs, slime}.

Online co-training keeps the draft model aligned with the policy model during RL~\citep{qian2026drafts}. Typically, offline drafts are trained on a fixed corpus with a fixed target model, whereas RL rollouts come from task-specific prompts and a policy that keeps changing~\citep{wang2026aurora}. Co-training the draft model on these rollouts improves acceptance length and rollout throughput~\citep{zhao2024slimetutorial,wang-etal-2026-mtp}. DeepSeek-V4.1-Flash~\citep{deepseekai2026deepseekv41flash} co-trains its DSpark~\citep{cheng2026dspark} draft during post-training. Co-training also lets the draft reuse the target features computed in the policy update instead of a separate target forward pass.

However, training large models on long contexts requires context parallelism (CP) and pipeline parallelism (PP) to fit activations and weights in GPU memory, yet co-training a draft under these parallelisms remains unexplored. Maintaining a separate parallelism strategy for the draft would add implementation complexity and per-step overhead for redistributing (sequences, target features, etc.) between two layouts. Therefore, draft co-training should operate directly on the policy's existing CP and PP layouts. As shown in Figure~\ref{fig:problem}, this poses two obstacles: (a) Advanced drafts such as EAGLE-3~\citep{li2026eagle}, DFlash~\citep{chen2026dflash}, and DSpark introduce branch-structured attention, where draft tokens branch off the main sequence at anchor positions and attend to both the main-sequence prefix and one another. Standard CP does not support this structure. (b) Under PP, the target features consumed by the draft model may reside on PP stages remote from it.

To address the two obstacles, we design and integrate online draft co-training system into NeMo-RL~\citep{nemo-rl}, with Megatron~\citep{megatron-lm} as training backend and vLLM~\citep{kwon2023efficient} as inference backend. Under CP, the draft shares the policy's packed zigzag-ring layout and scales to long contexts together with the policy. Under PP, target features are delivered outside the pipeline communication, leaving the schedule order unaffected.
For CP, we split the attention into two key sets: a causal prefix of the main sequence and the branch keys. We compute the causal component with packed zigzag-ring attention, handle the branch-local component on the owning rank, then merge the two. For PP, \emph{TapChannel} transports intermediate target features across stages into a mailbox on a side path independent of pipeline communication. 

Extensive experiments across draft families, model scales, and single- and multi-turn RL workloads demonstrate stable learning and scalable system performance. Co-training keeps the policy's reward and accuracy close to the baseline while delivering 1.19--1.88$\times$ end-to-end training speedups. Our CP implementation is up to
2.9$\times$ faster than USP (Unified Sequence Parallelism) with 2.7$\times$ lower per-GPU memory and scales to 256K tokens. TapChannel supports cross-stage feature delivery with modest transport overhead.

Our contributions are summarized as follows:
\begin{itemize}
\item \textbf{Branch attention under CP.} We decompose draft branch attention into a causal main-sequence component on the policy's packed, load-balanced zigzag ring and a rank-local branch component. This unified mechanism supports EAGLE-3, DFlash, and DSpark.
\item \textbf{Out-of-schedule target-feature transport under PP.} We introduce \emph{TapChannel}, which delivers intermediate target features across PP stages without altering the pipeline schedule.
\item \textbf{End-to-end co-training at scale.} We implement and evaluate the system in NeMo-RL on three draft families, 8B--122B targets, and single- and multi-turn RL workloads, covering learning stability, speedups, and overhead.
\end{itemize}

\input{tex_figure/problem}

%% file: tex_figure/problem.tex
\begin{figure*}[t]
  \centering
  \begin{minipage}[b]{0.60\textwidth}
    \centering
    \includegraphics[width=\linewidth]{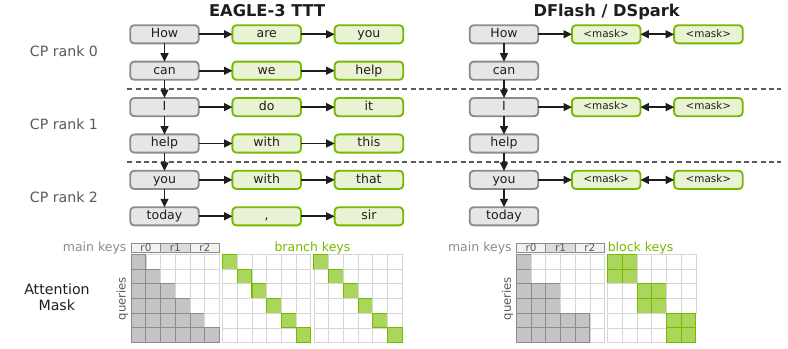}\\[3pt]
    {\small (a) CP: branch attention is not a causal or full mask.}
  \end{minipage}\hfill
  \begin{minipage}[b]{0.385\textwidth}
    \centering
    \includegraphics[width=\linewidth]{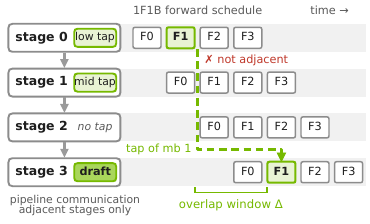}\\[3pt]
    {\small (b) PP: taps (target hidden states) come from non-adjacent pipeline stages.}
  \end{minipage}
  \caption{Two obstacles to online draft co-training. (a) The main sequence (gray) is split across CP ranks. Draft attaches branches (green) to it. (b) Target features (taps) are produced on stages far from the draft, but pipeline communication connects only adjacent stages.}
  \label{fig:problem}
\end{figure*}

%% file: sec/method.tex
\section{System Design and Implementation}
\label{sec:method}
\input{tex_figure/cp_branch_attention}

We first describe the online draft co-training procedure in RL post-training, and then present the two mechanisms that support it: branch attention under context parallelism (§\ref{subsec:cp}) and TapChannel under pipeline parallelism (§\ref{subsec:pp}).

\subsection{Online Draft Co-Training in RL Post-Training}
\label{subsec:setup}

Let $\theta$ and $\phi$ denote the policy and draft parameters. The draft is trained on the same policy rollout tokens, with a stop-gradient applied to the target features ($H_\theta$) collected from intermediate policy layers. The joint objective is
\begin{equation}
  \mathcal{L}(\theta,\phi)
  \;=\; \mathcal{L}_{\text{RL}}(\theta)
  \;+\; \lambda\,\mathcal{L}_{\text{draft}}
  \bigl(\phi; x, \operatorname{sg}(H_\theta(x))\bigr),
\end{equation}
where $x$ denotes the rollout samples and $\operatorname{sg}$ denotes stop-gradient operation.

We integrate the draft model co-training into the RL training loop. At each iteration, the policy and draft model generate rollouts through speculative decoding. During policy training, we capture target features from selected policy layers and use them to train the draft model on the same microbatch. The draft model resides on the last pipeline stage, where its loss is added to the policy loss for a shared backward pass and optimizer step. Under CP, the policy's packed sequence shard and locally captured features are shared with the draft model; under PP, TapChannel delivers target features from earlier stages.

\subsection{Branch Attention under Context Parallelism}
\label{subsec:cp}
\paragraph{Branch structure.} The attention mechanism in draft model attaches \emph{branches} to the \emph{main sequence}. A branch is a short run of draft queries that starts at an \emph{anchor} position. Each branch query attends to the main sequence up to its anchor and to the branch's own keys. The draft families differ in what a branch holds. In EAGLE-3 TTT~\citep{li2026eagle}, which simulates multi-step drafting by feeding the draft's own hidden states back as inputs, every position is an anchor, and its branch keys are the draft's hidden states from earlier TTT passes. DFlash~\citep{chen2026dflash} and DSpark~\citep{cheng2026dspark} use block diffusion to generate an entire block of tokens in a single forward pass, so a branch is a block of positions that attend to one another bidirectionally. Branch attention is therefore neither causal nor full, the two mask types that ring-attention CP kernels support, as illustrated in Figure~\ref{fig:problem}a.

\paragraph{Design choice.} SpecForge~\citep{li2026specforge} implements EAGLE-3 TTT via sequential ring sharding and Ulysses, which yields imbalanced causal workloads, and imposes constraints on the Ulysses dimension that can be restrictive for draft KV heads. We keep the policy's own packed zigzag-ring attention layout for the causal component and compute the branch component on the CP rank that owns its anchor. This preserves the existing training parallelism configuration and load balance. We then merge the two attention results to recover the output for each branch query.

\paragraph{Decomposition and merge.} The main-sequence component follows the same packed zigzag-ring attention as standard CP implementation. For each ring step, the main-sequence K/V circulate across ranks while queries remain local; the locally computed branch component is then merged via Eq.~\eqref{eq:lse-merge}. 
The merge follows the standard online-softmax reduction used between ring-attention steps:
\begin{equation}
  \ell = \log\!\left(e^{\ell_{\mathrm{m}}}
  + e^{\ell_{\mathrm{b}}}\right),
  \qquad
  O = e^{\ell_{\mathrm{m}}-\ell}\,O_{\mathrm{m}}
  \;+\; e^{\ell_{\mathrm{b}}-\ell}\,O_{\mathrm{b}},
  \label{eq:lse-merge}
\end{equation}
where $(O_{\mathrm{m}},\ell_{\mathrm{m}})$ and $(O_{\mathrm{b}},\ell_{\mathrm{b}})$ correspond to main-sequence and branch. $O$ denotes the attention output and $\ell$ denotes the log-sum-exp for the main-sequence and branch-local components. Since all three drafts differ only in their branch-local keys, they share the same decomposition.

\paragraph{Communication cost and overlap.}
For a CP degree \( C \) and \( N \) main-sequence tokens evenly split across ranks, each rank holds \( N/C \) tokens. During the forward ring, each rank sends its local K and V to the other \( C-1 \) ranks, yielding a per-rank outbound volume of
\begin{equation}
  V_{\mathrm{CP}}^{\mathrm{fwd}}
  = 2(C-1)\frac{N}{C}d_{\mathrm{kv}}b,
  \label{eq:cp-comm}
\end{equation}
where \( d_{\mathrm{kv}} \) is the per-token K/V width and \( b \) the bytes per element. Since branch-local K/V stay on their anchor-owning ranks, this cost is independent of the number or depth of branches. The backward pass replays the same K/V ring and accumulates gradients locally, incurring no additional branch-dependent traffic.

We overlap communication with attention at each ring step: we initiate the next shard transfer before computing attention on the current shard, then wait for the transfer to complete before processing the next shard. Let \( t_{\mathrm{comm}}^{(r)} \) and \( t_{\mathrm{attn}}^{(r)} \) denote the communication and attention durations at step \( r \). The elapsed time for their overlapped execution is
\begin{equation}
  E_{\mathrm{CP}}^{(r)}
  = \max\left(t_{\mathrm{comm}}^{(r)}, \; t_{\mathrm{attn}}^{(r)}\right).
  \label{eq:cp-overlap}
\end{equation}
The slower operation determines the ring-step duration: attention dominates when the exchange completes within the computation interval, and communication dominates otherwise. Since attention computation scales quadratically with context length while communication scales linearly, longer contexts increase the overlap headroom. Conversely, aggressive strong scaling shortens the local context and can eventually expose communication as the bottleneck.

\subsection{TapChannel under Pipeline Parallelism}
\label{subsec:pp}
\input{tex_figure/tap_channel}
\paragraph{Taps on non-adjacent stages.} Under pipeline parallelism, the layers producing the taps (target hidden states) span several stages, while the draft resides on the last stage. Standard pipeline communication connects adjacent stages, so forwarding a tap from a non-adjacent source would add send and receive operations to the intervening stages' 1F1B schedules. TapChannel instead transfers each tap directly to the draft stage through a separate path, as illustrated in the left panel of Figure~\ref{fig:tap-channel}. Since the features are detached during draft training, these transfers require no backward communication. The interval between the source forward pass and the draft stage's forward pass allows feature transport to overlap with the policy model's pipeline computation, including work on other microbatches.

\paragraph{Alternatives considered.}
TapChannel implements this separate path using RDMA one-sided writes. We considered three alternatives to TapChannel. (1) Adding target features to existing pipeline messages would increase message size and require intermediate stages to receive and forward features they do not use. (2) Inserting two-sided sends and receives directly into the pipeline schedule would require careful coordination: a source stage may reach a send before the draft stage reaches the corresponding receive, and waiting there could stall the pipeline. (3) Extracting features in inference engines could reduce rollout throughput, while using a separate target-model instance would require additional resources. These considerations motivated TapChannel's direct transfers from source stages to the draft stage, using one-sided writes. We also implemented pinned-host-memory staging as a transport baseline, but its additional GPU--CPU transfers make fan-in 4.5--8.5$\times$ slower than TapChannel and slow computation across all stages by 83--88\%.

\paragraph{Per-source mailbox.} TapChannel implements this path with a per-source mailbox in the draft stage. As shown by the right panel of Figure~\ref{fig:tap-channel}, each mailbox is a ring of pre-allocated slots: the source writes the target features for microbatch $m$ into slot $m \bmod S$, and the draft reads it before the forward pass for $m$. Each slot carries a stamp so that the draft can verify that the features belong to the corresponding microbatch. The source never overwrites a slot whose tap is still in use: it writes a new tap into a slot only after the draft has finished reading the old one.

Within a node, TapChannel uses CUDA IPC, and the draft clears the stamp after reading to release the slot. Across nodes, TapChannel transfers features through a dedicated NCCL communicator using GPUDirect RDMA. Here, stamps track operation order rather than signal that a slot is free: the sender limits outstanding sends and waits for completion before reusing send buffers, while CUDA events on the receiver prevent incoming transfers from overwriting features which is still being read by the draft.

The mailbox must hold features that the source has produced but the draft has not yet consumed, while leaving room for delays in slot release. For source stage $r$ in a $P$-stage pipeline, $P-r$ bounds the number of in-flight microbatches under our non-interleaved 1F1B schedule. With exactly $P-r$ slots, however, the source may revisit a slot just as the draft releases it. Even a short delay before the source observes the release can stall its forward pass, leaving downstream stages waiting for activations. We therefore default to $S=2(P-r)$ slots to give each slot more time to become available before the source needs it again.

\paragraph{Communication cost and overlap.}
Let \( n \) be the number of token rows in a microbatch and \( \mathcal{S} \) the set of source stages that send features to the draft stage. Let \( d_s \) be the per-token feature dimension produced by source \( s \). For the first stage, \( d_s \) equals the input embedding width; for later stages, it equals the hidden size. Let \( b \) be the bytes per element. The feature payload delivered per microbatch is
\begin{equation}
  V_{\mathrm{Tap}} = bn\sum_{s\in\mathcal{S}} d_s.
  \label{eq:tap-comm}
\end{equation}
Each feature is transferred directly once, regardless of the number of PP hops between its source and the draft; features produced on the draft stage itself require no transfer.

For microbatch \(m\), let \(\tau_{s,m}\) be the time required to transfer target features from source stage \(s\) to the draft stage, and let \(\Delta_{s,m}\) be the time from feature production to the draft's scheduled forward pass start. The longer of these two intervals therefore determines when that source's features can be used. Taking the maximum across sources gives
\begin{equation}
  E_{\mathrm{Tap}}^{(m)}
  = \max_{s\in\mathcal{S}} \max\left(\tau_{s,m}, \; \Delta_{s,m}\right),
  \label{eq:tap-overlap}
\end{equation}
where each interval is measured from feature production on its source stage. If all transfers finish before the scheduled draft forward pass, the draft incurs no additional wait. Otherwise, it waits for the remaining target features. Thus, \(E_{\mathrm{Tap}}^{(m)}\) measures the longest interval from feature production on a source stage until both its transfer is complete and the draft is scheduled to run.

%% file: tex_figure/cp_branch_attention.tex
\begin{figure*}[t]
  \centering
  \includegraphics[width=\linewidth]{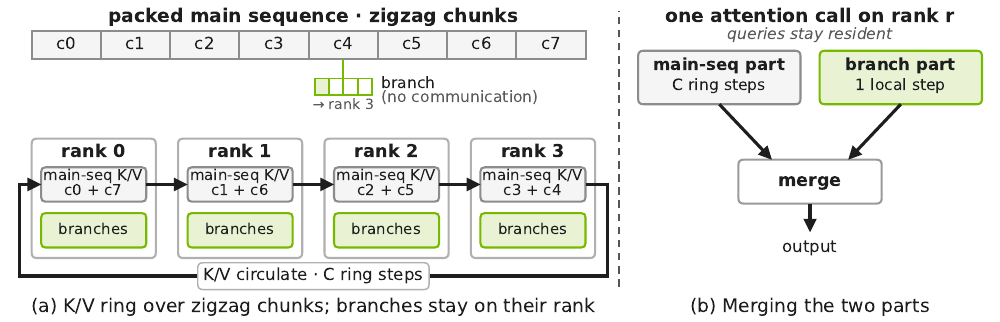}
  \vspace{-1em}
  \caption{Branch attention under context parallelism (CP).
(a) The prefix sequence is divided into chunks, with each rank assigned
an early and a late chunk (zigzag sharding) to balance causal attention
computation. Queries remain on their assigned ranks, while prefix keys
and values circulate around the ring so that each query can attend to
its allowed prefix. Branch keys and values stay on the same rank as
their queries and require no communication.
(b) Each rank combines prefix attention from the ring with local
branch attention using log-sum-exp weights
(Eq.~\eqref{eq:lse-merge}), producing the same result as attention
over the combined visible keys.}
  \label{fig:cp-branch-attention}
\end{figure*}

%% file: tex_figure/tap_channel.tex
\begin{figure*}[t]
  \centering
  \includegraphics[width=\linewidth]{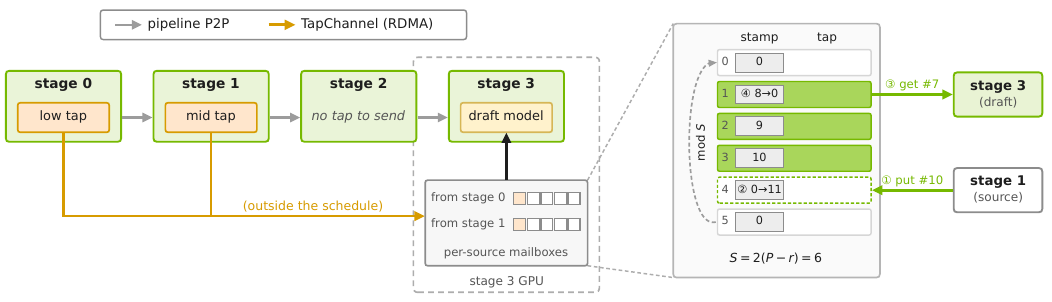}
  \vspace{-1em}
  \caption{Illustration of TapChannel. Left: pipeline activations follow the
scheduled P2P path (gray arrow), while stages that own taps send them over a
separate path (orange) into per-source mailboxes on the draft stage. Right:
the mailbox of stage 1, a ring of $S=6$ slots, each holding a tap and a
stamp. (1) The source writes the $n$-th tap into
slot $n \bmod S$ and (2) sets its stamp to
$n{+}1$; (3) the draft reads the tap once the
stamp matches and (4) resets the stamp to free
the slot.}
  \label{fig:tap-channel}
\end{figure*}

%% file: sec/experiments.tex
\section{Evaluation}
\label{sec:experiments}
We evaluate four questions: (1) whether draft co-training affects the RL learning process; (2) how much it accelerates end-to-end training across draft families, model scales, and single- and multi-turn workloads, and what bounds this speedup; (3) the scalability of branch attention under CP at long contexts; and (4) whether TapChannel's feature transfers can be overlapped with pipeline computation, and how the draft's compute affects pipeline stage balance.

\subsection{Experimental Setup}
\label{subsec:exp-setup}

\paragraph{Models.}
We evaluate three representative draft families: EAGLE-3~\citep{li2026eagle}, DFlash~\citep{chen2026dflash}, and DSpark~\citep{cheng2026dspark}, all with Qwen3-8B~\citep{qwen3} as the target. To test the
scalability and generality, we also evaluate larger models: Qwen3.5-27B, Qwen3.5-35B-A3B, Qwen3.5-122B-A10B~\citep{qwen35}, and GPT-OSS-120B with DFlash; Qwen3-14B and Nemotron-3.5-Lightning-30B-A3B~\citep{nvidia_nemotron_3_super} with DSpark. All drafts are initialized from official checkpoints.
\input{tex_figure/qwen3_8b_learning_and_performance}

\paragraph{Tasks and training.}
We conduct experiments in NeMo-RL~\citep{nemo-rl} with GRPO~\citep{shao2024deepseekmath} on DAPOMath-17K~\citep{yu2025dapo} and evaluate on AIME~2024~\citep{aime24}, with 4k input and 16k output using
vLLM~\citep{kwon2023efficient} inference engine colocated with training. Each policy update draws 512 responses (32 prompts $\times$ 16), or 256 (32 $\times$ 8) for Qwen3.5-122B-A10B and GPT-OSS-120B. On H100, Qwen3-8B trains with TP2, Qwen3-14B trains with TP4, and Qwen3.5-35B-A3B trains with TP/PP/CP/EP $=$ 2/2/2/8. On GB200, Qwen3.5-27B trains with TP/PP/CP $=$ 4/2/2, Nemotron-3.5-Lightning-30B-A3B trains with TP/PP/CP/EP $=$ 2/2/2/8, and Qwen3.5-122B-A10B and GPT-OSS-120B train with TP/PP/CP/EP $=$ 4/4/2/16. For multi-turn evaluation, we adapt the NeMo Gym Workplace Assistant~\citep{styles2024workbench} on Qwen3-8B with sequence length of 32{,}768 with tool feedback. Workplace Assistant is a multi-turn, agentic tool-use environment within NVIDIA NeMo Gym, designed for complex task execution in simulated office scenarios. It is based on the WorkBench benchmark~\citep{styles2024workbench}, a simulated workplace tool-use environment with email, calendar, CRM, project management, and analytics databases. Detailed setup can be found in Appendix~\ref{app:exp-config}.

\paragraph{Metrics.}
We report training reward, validation accuracy, and training-inference KL divergence: the per-token KL (K3 estimator) between the log-probabilities from the training and inference backends under identical weights, where low values indicate numerical consistency. For speculative decoding, we report acceptance length, the mean number of tokens accepted per verification. We also report rollout throughput, rollout speedup relative to the baseline, and end-to-end speedup in total policy-update time.
\input{tex_figure/qwen3_8b_workplace_learning_and_performance}
\input{tex_table/scale_table}
\input{tex_figure/step_cost}

\subsection{Verifying Policy Learning under Speculative Decoding}
\label{subsec:exp-stability}

We select Qwen3-8B as the target for all three draft models. We compare four settings: a baseline with neither online draft co-training nor rollout speculative decoding, and online co-training with EAGLE-3, DFlash, and DSpark. As shown in Figure~\ref{fig:learning-stability}, the reward, validation accuracy, and training--inference KL divergence confirm that speculative decoding preserves the RL learning trajectory. The co-trained runs closely track the baseline across all three draft models.

\subsection{Performance across Drafts and Model Scales}
\label{subsec:exp-main-performance}

Table~\ref{tab:main-performance} shows results across three draft families and targets from 8B to 122B. Co-trained drafts reach acceptance lengths of 2.28--5.10, yielding 1.48--2.25$\times$ rollout speedup and 1.19--1.88$\times$ end-to-end training speedup. Comparing the three draft models, DFlash and DSpark consistently reach longer acceptance lengths than EAGLE-3, which suggests larger speedups.
Notably, while the larger MoE targets (Qwen3.5-122B and GPT-OSS-120B) achieve long acceptance lengths, their end-to-end speedups are lower, since verifying several tokens in one forward pass activates more experts~\citep{huang2026moesd}. For linear-attention targets, verification is comparatively cheap, so the relative speedup from speculation is smaller~\citep{wang2026specla}.
\input{tex_figure/pack_dist}

\paragraph{Multi-Turn Workloads.}
\label{subsec:exp-multiturn}
Workplace Assistant interleaves multiple turns with tool calls and environment delays. As shown in Figure~\ref{fig:workplace-learning}, all settings achieve consistent rewards, while acceptance length improves across training. Note that the end-to-end speedup (1.26--1.41$\times$) falls below the rollout speedup (1.8--2.4$\times$), because rollout accounts for only 56\% of the step time, and tool execution and environment latency sit inside this phase, unreachable by faster decoding. In multi-turn workloads, the end-to-end gain is therefore bounded by the share of the step spent decoding.

\paragraph{Training-step time breakdown.}
Figure~\ref{fig:step-cost} breaks down the RL step time for Qwen3-8B with DSpark, Qwen3.5-27B with DFlash and Qwen3.5-122B-A10B with DFlash. Rollout dominates the step, taking 56--73\% of time, and speculative decoding shortens it by 1.3--2.4$\times$. Co-training adds 6--20\% to update time and at most 7\% to refit time, and leaves the remaining phases largely unchanged. Because rollout takes most of the step, its savings outweigh these added costs.

\input{tex_figure/cp_scaling}

\subsection{Context-Parallel Attention Performance}
\label{subsec:exp-cp}
We compare our packed zigzag-ring attention with the USP implementation in SpecForge~\citep{li2026specforge} at the attention-operator level, using the same number of GPUs. The benchmark uses EAGLE-3 with three TTT passes, the draft that SpecForge's USP supports. To our knowledge, no open-source implementation supports CP training for masked prediction models like DFlash or DSpark. We evaluate the two through scaling property instead. We report forward/backward latency and per-GPU peak HBM.

\paragraph{Analysis.} Figure~\ref{fig:pack-dist} compares our packed, zigzag-ring attention with USP on the same variable-length workload. The longest sequence is set to 20,480. USP pads the batch to 2.25$\times$ its real token count, whereas our packed zigzag layout processes only real tokens. At CP$=$2, 4, and 8, packed zigzag-ring attention is 2.9$\times$, 2.3$\times$, and 1.5$\times$ faster than the best USP variant, and it reduces peak per-GPU memory by 2.7$\times$. The speedup can be attributed mainly to skipping padding. As CP size grows, each GPU holds fewer tokens, and fixed per-step costs such as ring communication take up a larger share of the runtime, narrowing the gap.

\paragraph{Long-context scaling.}
We evaluate our implementation across CP$\in\{1,2,4,8\}$ at fixed global lengths from 16K to 256K tokens, as shown in Figure~\ref{fig:cp-scaling}. For EAGLE-3 TTT attention, latency drops from 17.7 s at CP=1 to 2.35 s at CP=8 (7.5$\times$, 94\% parallel efficiency), and per-GPU memory falls nearly linearly from 53.2 GB to 7.5 GB. Block-draft attention is launch-bound rather than FLOP-bound, which works in its favor: it stays cheap even at CP=8 ($\le$55 ms from 16K to 256K), yet scales 6.9$\times$ at 256K when the workload is substantial. Context parallelism thus rides the policy's layout nearly for free.

\input{tex_figure/tap_transport}

\subsection{Pipeline-Parallel Overhead}
We measure the incremental cost of adding TapChannel to an existing PP schedule, comparing the full method against the same pipeline-parallel policy with draft training disabled. We compare against a host-staging baseline that stages taps through pinned host memory.

\paragraph{Transport microbenchmark.}
We first measure TapChannel's transport on H100 GPUs, with the fan-in and the per-stage payloads taken from EAGLE-3 and DFlash at PP$=4$: two or three source stages deliver 32--256\,MB each, 64--512\,MB per fan-in, at 4K- and 16K-token microbatches. As shown in Figure~\ref{fig:tap-transport}, one-sided writes achieve 27--39\,GB/s and complete the fan-in 4.5--8.5$\times$ faster than staging through pinned host memory. Side-stream transfers have negligible impact on source-stage computation and increase computation time on the receiving draft stage by only 1.6\%, whereas host staging slows computation by 83--88\% across all stages.
\input{tex_figure/pp_balance}
\input{tex_table/tap_channel_table}

\paragraph{TapChannel overhead in RL training.}
Table~\ref{tab:tap-channel-runs} reports the feature payload per microbatch, exposed communication wait, and mailbox memory in the RL runs. Payload is computed using Eq.~\eqref{eq:tap-comm} at the allocated slot capacity, matching the amount transferred by the cross-node path. Exposed wait measures how long the draft stage waits for required features. On Qwen3-8B, this wait accounts for 1.5--4.9\% of update time across PP$=2$ and PP$=4$. On Qwen3.5-35B-A3B, where two of the three source stages send across nodes, it totals 2.0\,s per update (2.2--2.7\%). On GPT-OSS-120B and Qwen3.5-122B-A10B, all three source stages send across nodes, yet the wait stays below 0.5\% of update time (0.31--0.63\,s per update). Mailbox memory is the total memory reserved for feature buffers on the draft stage. On Qwen3-8B, it increases from 2.1--2.8\,GB at PP$=2$ to 3.9--5.6\,GB at PP$=4$; the three larger models at PP$=4$ require 2.5--3.4\,GB, as their per-microbatch payloads are smaller (396--570\,MB versus 528--881\,MB). TapChannel's transfers are thus largely overlapped with pipeline computation; the remaining cost of PP co-training is the draft's compute on the last stage.

\subsection{Balancing Pipeline Stages for the Draft}
\label{subsec:exp-pp-balance}

The draft runs entirely on the last pipeline stage. Deeper pipelines spread the policy layers over more stages, but the draft's cost stays on that one stage, so the last stage becomes increasingly heavier than the rest. Because the slowest stage sets the pipeline period, the other stages then idle while they wait for it. We therefore move some policy layers from the last stage to earlier stages, choosing how many to move from measured per-stage times.

\paragraph{Setup.}
We study Qwen3.5-35B-A3B (40 layers) with PP=4 on H100; the draft stage and two of the three source stages sit on different nodes. Every stage carries per-microbatch timers around its forward and backward, from which we read the per-stage compute and the steady-state period. We compare the even layer split with a split that moves layers off the last stage, 11/11/11/7.

\paragraph{Sources of stage imbalance.}
Without a draft model, the last stage is already the slowest (shown by Figure~\ref{fig:pp-balance}, even split):
besides its Transformer layers, it computes the vocabulary projection and the loss. Co-training
adds about 120\,ms of forward and backward compute per microbatch to this stage and lengthens the
pipeline period by 127\,ms, so the draft accounts for nearly all of the added time. Feature
transfers also slightly slow the source stages, but they remain faster than the last stage and do not set
the period.

\paragraph{Balancing by measured stage time.}
We choose the split from the measured stage times, counting both the output head and the draft
toward the last stage. On Qwen3.5-35B-A3B, moving three layers off the last stage (11/11/11/7)
shortens the pipeline period to 622\,ms, and a co-training update (73.3\,s) then runs faster than a no-draft update under the even split (76.3\,s). Part of this gain comes from balancing the policy itself: the same split lowers the no-draft update time from 76.3 to 65.2\,s, so under a matched layout co-training adds 12\% to update time.

%% file: tex_figure/qwen3_8b_learning_and_performance.tex
\begin{figure*}[t]
  \centering
  \includegraphics[width=\linewidth]{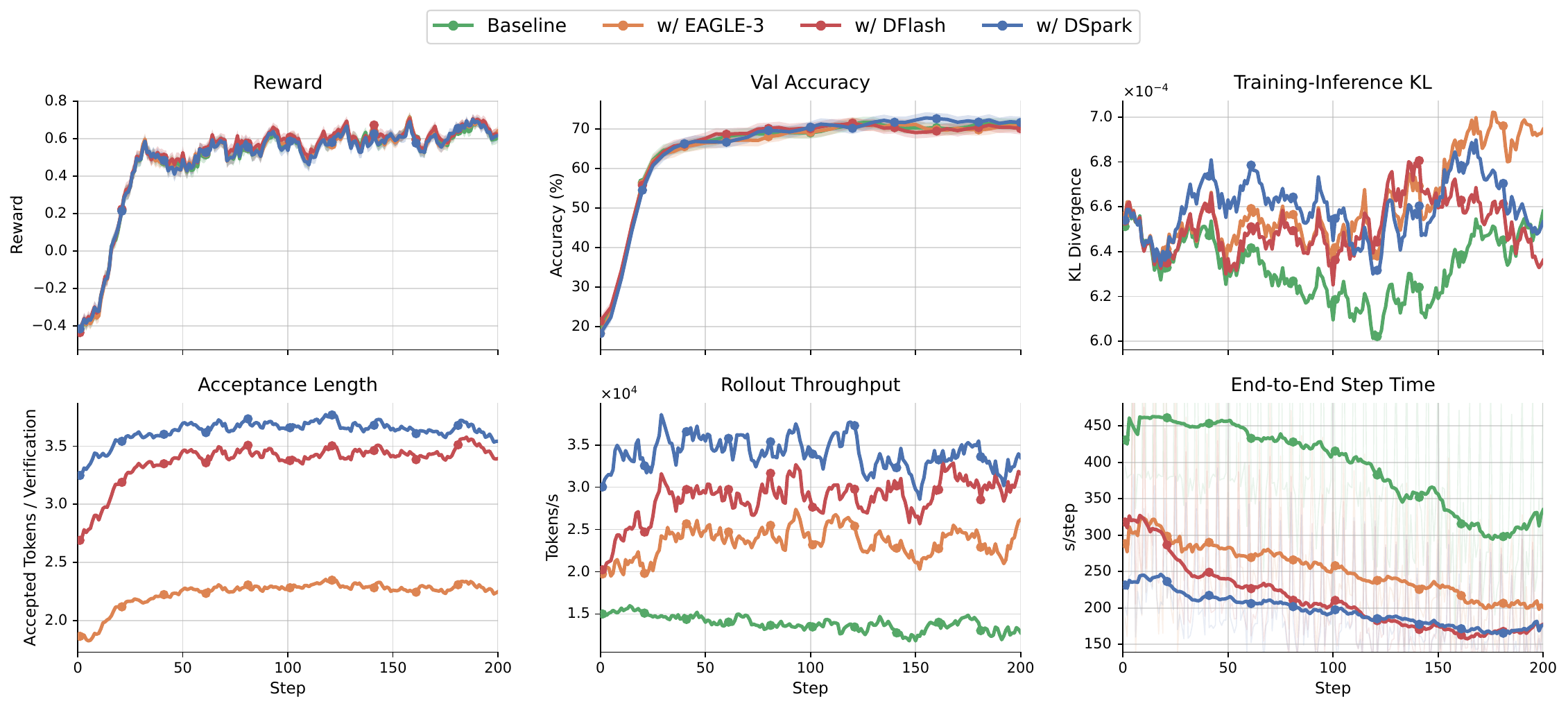}
  \vspace{-2em}
  \caption{Experimental results of Qwen3-8B online co-training with EAGLE-3 (yellow), DFlash (red) and DSpark (blue) on DAPOMath-17K, including reward, accuracy evaluated on AIME2024, KL divergence, accepted length, throughput and end-to-end step time.}
  \label{fig:learning-stability}
\end{figure*}

%% file: tex_figure/qwen3_8b_workplace_learning_and_performance.tex
\begin{figure*}[t]
  \centering
  \includegraphics[width=\linewidth]{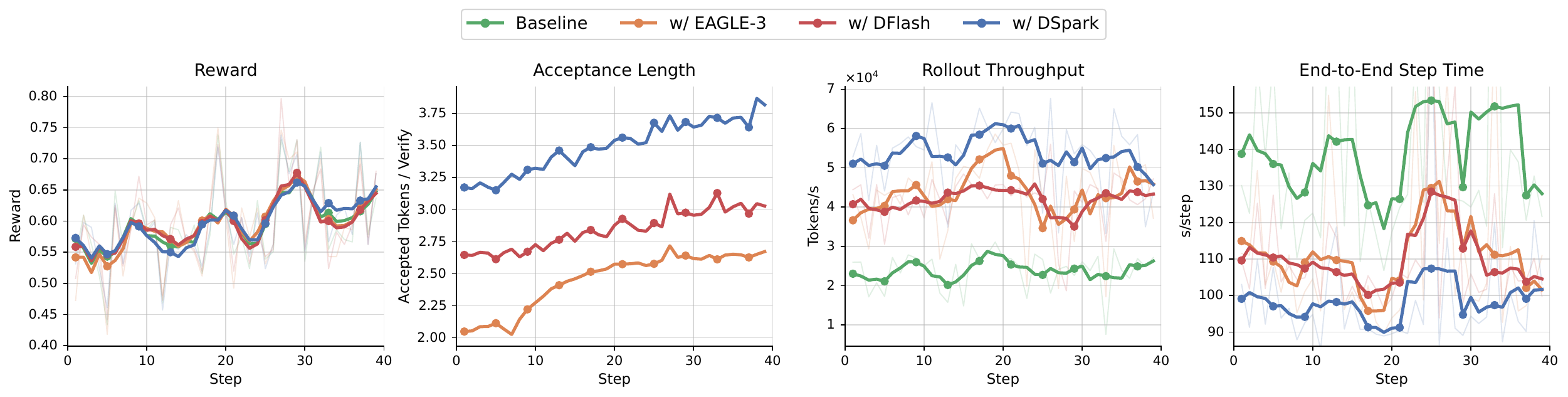}
  \vspace{-2em}
  \caption{Qwen3-8B online co-training with EAGLE-3 (yellow), DFlash (red), and DSpark (blue) on NeMo Gym Workplace Assistant.}
  \label{fig:workplace-learning}
\end{figure*}

%% file: tex_table/scale_table.tex

\begin{table*}[t]
  \centering
  \caption{End-to-end performance across target scales and draft families.}
  \resizebox{\textwidth}{!}{%
  \begin{tabular}{llccc}
    \toprule
    Target Model & Draft Model & Accepted Length $\uparrow$ & Rollout Speedup $\uparrow$ & E2E Training Speedup $\uparrow$ \\
    \midrule
    \multirow{3}{*}{Qwen3-8B} & EAGLE-3 & 2.28 & 1.63$\times$ & 1.50$\times$ \\
    & DFlash & 3.45 & 2.23$\times$ & 1.88$\times$ \\
    & DSpark & 3.63 & 2.18$\times$ & 1.83$\times$ \\
    Qwen3-14B & DSpark & 3.61 & 2.25$\times$ & 1.78$\times$ \\
    \midrule
    Qwen3.5-27B & DFlash & 5.10 & 1.52$\times$ & 1.26$\times$ \\
    Qwen3.5-35B-A3B & DFlash & 4.58 & 1.50$\times$ & 1.46$\times$ \\
    Nemotron-3.5-Lightning-30B-A3B & DSpark & 3.84 & 1.67$\times$ & 1.46$\times$ \\
    \midrule
    Qwen3.5-122B-A10B & DFlash & 4.90 & 1.72$\times$ & 1.43$\times$ \\
    GPT-OSS-120B & DFlash & 3.80 & 1.48$\times$ & 1.19$\times$ \\
    \bottomrule
  \end{tabular}}%
  \label{tab:main-performance}
\end{table*}

%% file: tex_figure/step_cost.tex
\begin{figure*}[t]
  \centering
  \includegraphics[width=\linewidth]{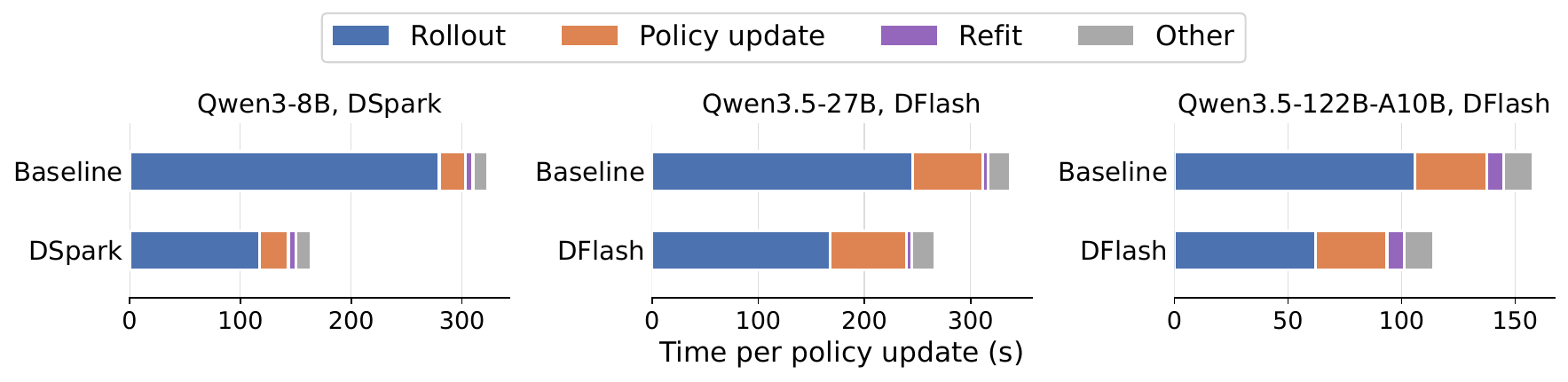}
  \vspace{-2em}
  \caption{Time per policy update by phase, without and with draft co-training.  Other includes data preparation, reward
  and advantage processing, and checkpointing.}
  \label{fig:step-cost}
\end{figure*}

%% file: tex_figure/pack_dist.tex
\begin{figure*}[t]
  \centering
  \includegraphics[width=\linewidth]{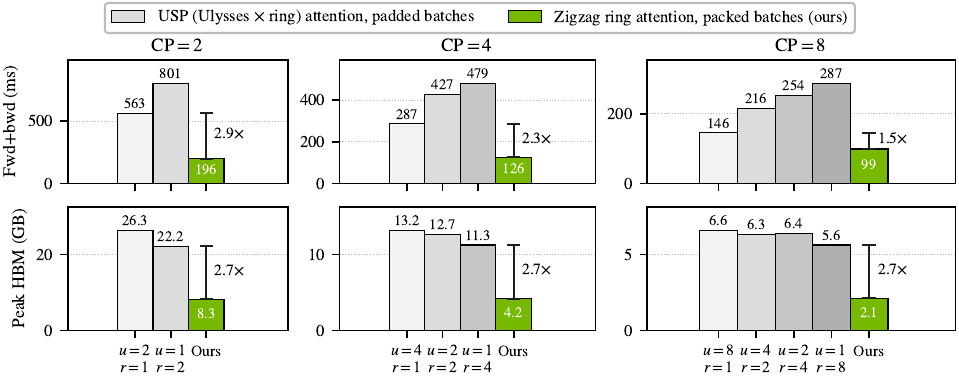}
  \vspace{-1.5em}
  \caption{Comparison of packed zigzag-ring attention (ours) vs. padded USP attention under the same workload. Top: latency of the forward+backward pass; bottom: per-GPU peak memory. Each bracket marks the advantage over the best USP configuration.}
  \label{fig:pack-dist}
\end{figure*}

%% file: tex_figure/cp_scaling.tex
\begin{figure*}[h]
  \centering
  \includegraphics[width=\linewidth]{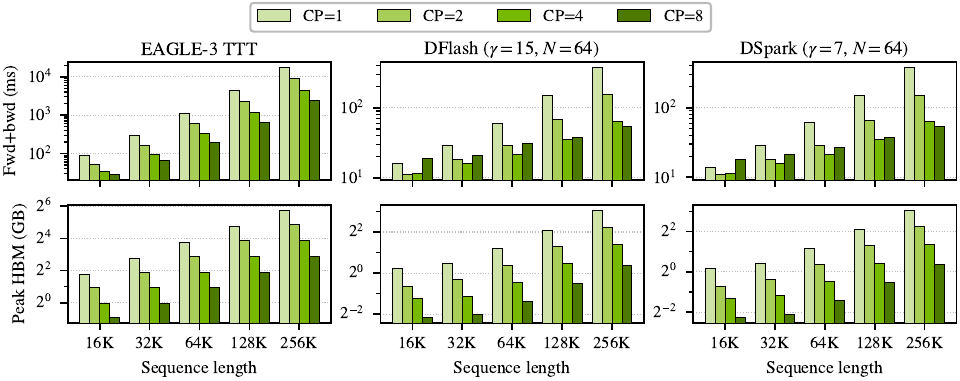}
  \vspace{-1em}
  \caption{Scaling of draft-training attention over CP=1, 2, 4, 8. Each group fixes the global sequence length and reports forward+backward latency (top) and peak HBM (bottom).}
  \label{fig:cp-scaling}
\end{figure*}

%% file: tex_figure/tap_transport.tex
\begin{figure}[t]
  \centering
  \includegraphics[width=0.75\linewidth]{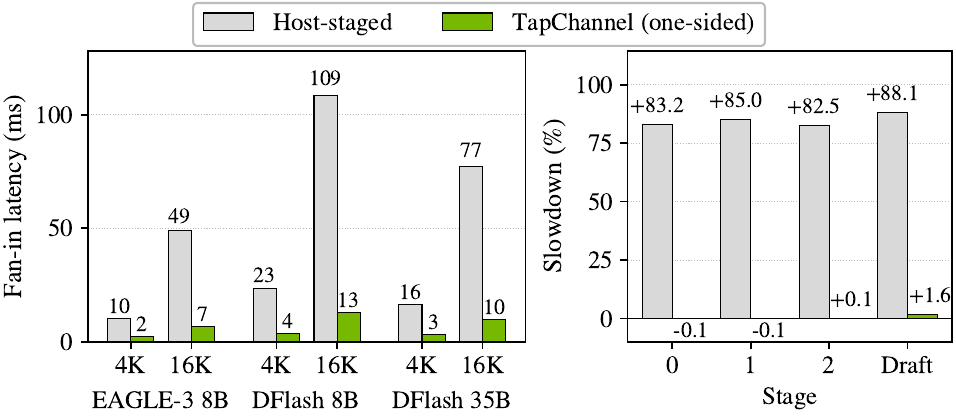}
  \vspace{-1em}
  \caption{TapChannel vs. host staging. Left: fan-in latency; right: slowdown during transfers, measured as the time increased when transfers run concurrently. One-sided writes complete 4.5–8.5× faster than host staging, and cost the receiving draft stage only 1.6\% in HBM contention; host staging slows every rank by 83–88\%.}
  \label{fig:tap-transport}
\end{figure}

%% file: tex_figure/pp_balance.tex
\begin{figure*}[t]
  \centering
  \includegraphics[width=\linewidth]{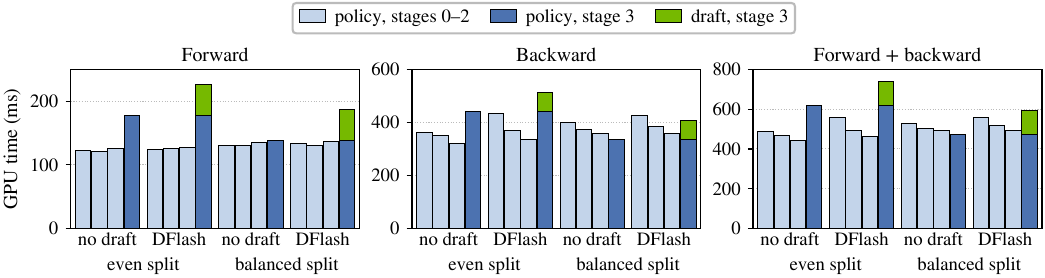}
  \vspace{-2em}
  \caption{Pipeline stage balance for Qwen3.5-35B-A3B at PP$=4$. Bars show per-microbatch time for stages 0--3. The darker bar marks the last stage. Green segments show the last-stage draft overhead. Balanced split: 11/11/11/7.}
  \label{fig:pp-balance}
\end{figure*}

%% file: tex_table/tap_channel_table.tex
\begin{table*}[t]
  \caption{TapChannel communication and memory costs. Source stages: stages sending target features, total / cross-node. Payload: feature volume per microbatch. Mailbox: buffer memory reserved on the draft stage. Tap wait: waiting time per policy update and its share of update time. Balanced splits: 11/11/11/7 (Qwen3.5-35B-A3B), 10/11/11/4 (GPT-OSS-120B), and 13/13/13/9 (Qwen3.5-122B-A10B).}
  \label{tab:tap-channel-runs}
  \centering
  \small
  \setlength{\tabcolsep}{7pt}
  \renewcommand{\arraystretch}{1.05}
  \begin{tabular}{@{}lccccc@{}}
    \toprule
    \multirow{2}{*}{Draft}
    & \multirow{2}{*}{\shortstack{Source stages\\Total / cross-node}}
    & \multirow{2}{*}{\shortstack{Payload\\(MB/mb)}}
    & \multirow{2}{*}{\shortstack{Mailbox\\(GB)}}
    & \multicolumn{2}{c}{Tap wait} \\
    \cmidrule(lr){5-6}
    & & & & Time (s) & Update (\%) \\
    \midrule
    \multicolumn{6}{l}{\emph{Qwen3-8B, PP$=2$}} \\
    EAGLE-3 & 1 / 1 & 528 & 2.11 & 0.62 & 1.5 \\
    DFlash  & 1 / 1 & 705 & 2.82 & 0.80 & 2.3 \\
    DSpark  & 1 / 1 & 705 & 2.82 & 0.82 & 2.4 \\
    \midrule
    \multicolumn{6}{l}{\emph{Qwen3-8B, PP$=4$}} \\
    EAGLE-3 & 2 / 2 & 528 & 3.88 & 1.25 & 2.4 \\
    DFlash  & 3 / 3 & 881 & 5.64 & 2.05 & 4.8 \\
    DSpark  & 3 / 3 & 881 & 5.64 & 2.05 & 4.9 \\
    \midrule
    \multicolumn{6}{l}{\emph{Qwen3.5-35B-A3B, PP$=4$}} \\
    DFlash, even split     & 3 / 2 & 499 & 3.14 & 2.00 & 2.2 \\
    DFlash, balanced split & 3 / 2 & 570 & 3.42 & 2.00 & 2.7 \\
    \midrule
    \multicolumn{6}{l}{\emph{GPT-OSS-120B, PP$=4$}} \\
    DFlash, even split     & 3 / 3 & 501 & 3.21 & 0.62 & 0.36 \\
    DFlash, balanced split & 3 / 3 & 501 & 3.41 & 0.63 & 0.47 \\
    \midrule
    \multicolumn{6}{l}{\emph{Qwen3.5-122B-A10B, PP$=4$}} \\
    DFlash, even split     & 3 / 3 & 396 & 2.49 & 0.31 & 0.37 \\
    DFlash, balanced split & 3 / 3 & 396 & 2.49 & 0.32 & 0.42 \\
    \bottomrule
  \end{tabular}
\end{table*}

%% file: sec/related_work.tex
\section{Related Work}
\label{sec:related_work}

\paragraph{Speculative decoding for RL rollouts.}
Speculative decoding accelerates generation by letting a lightweight draft propose multiple tokens for parallel verification by the target model, with rejection sampling preserving the target's output distribution. \citep{leviathan2023fast,chen2023accelerating}. Recent systems adapt speculative decoding to the rollout engine in different ways. NeMo-RL~\citep{nemo-rl} integrate MTP (Multi-Token Prediction) and EAGLE-3 draft models into both synchronous and asynchronous RL rollouts and study how deployment and draft configurations affect the end-to-end training speed \citep{iso2026accelerating}. SPEC-RL~\citep{liu2025spec} avoids a learned drafter altogether: it reuses response segments from the preceding iteration as speculative prefixes and verifies them under the current policy. EfficientRollout~\citep{kim2026efficientrollout} constructs a quantized self-drafter from the target, adjusts speculative length according to acceptance, and enables speculation only when the runtime is likely to benefit. Distribution-aware speculative decoding~\citep{shao2026beat} models the long-tail token distribution encountered during RL, improving draft acceptance as the policy distribution shifts. These works establish that speculative decoding can accelerate RL in practice, but primarily optimize draft construction and verification on the rollout side. We complement these efforts by integrating online draft co-training into distributed training. 

\paragraph{Draft adaptation under evolving RL policies.}
A fixed draft loses alignment with an evolving policy; adapting it during RL recovers acceptance length and speedup. FastGRPO combines online draft learning with concurrency-aware configuration~\citep{zhang2026fastgrpo}; ReSpec dynamically selects speculative parameters and distills the policy into the drafter~\citep{chen2026respec}. Parallel work adapts MTP modules: MTP-RL uses a parameter-sharing MTP module with advantage-aware optimization~\citep{wang-etal-2026-mtp}; OCC derives an adaptive coefficient to balance auxiliary loss against policy update~\citep{wang2026joint}; Bebop trains using total-variation objectives and studies pre-RL adaptation~\citep{li2026breaking}. From an online-learning perspective, draft co-training~\citep{qian2026drafts} is formulated as an adaptation problem, allowing the draft model to keep pace with the updating target policy instead of remaining fixed. These methods improve draft adaptation at the objective level; we address the systems challenge of making co-training practical under CP and PP.

\paragraph{Target-feature-conditioned and parallel draft architectures.}
Draft architectures trade proposal quality for sequential cost. Blockwise parallel decoding introduced multiple future-token predictors with prefix validation~\citep{stern2018blockwise}. MTP generalizes this as an auxiliary objective~\citep{gloeckle2024metamtp}; FastMTP reduces cost with a shared head and recursive conditioning~\citep{cai2025fastmtp}. EAGLE-3 conditions a separate drafter on fused target-layer features and uses TTT (Train-Time Test) to expose multi-step errors~\citep{li2026eagle}. DFlash produces candidate blocks in parallel via block diffusion~\citep{chen2026dflash}; DSpark combines a parallel backbone with a Markov head for dependency restoration~\citep{cheng2026dspark}; Domino similarly separates parallel modeling from sequential dependency~\citep{huang2026domino}. These architectures differ in training branches, attention masks, and feature interfaces.

Integrating these draft models into current distributed training systems remains a challenge. Large-model training combines tensor, pipeline, and other parallelism~\citep{megatron-lm,yan2026scalable}. As for long-context training, RingAttention distributes long sequences and overlaps KV communication with computation~\citep{liu2024ringattention}. These methods support causal attention training but do not handle draft-specific branch attention or cross-stage feature routing. P-EAGLE parallelizes EAGLE with structured masks~\citep{hui2026p}; LongSpec addresses long-context inference with bounded KV cache and positional adaptation~\citep{yang2026longspec}. SpecForge develops target-draft decoupling and hybrid-parallel TTT training, including sequence-parallel branch attention~\citep{li2026specforge}. Our setting differs from these works. The target is an evolving RL policy trained with existing CP and PP parallelism. Unlike previous work that modifies the parallel layout to accommodate draft training, we keep the target's topology unchanged and instead adapt the system around it.

%% file: sec/conclusion.tex



\section{Conclusion}
\label{sec:conclusion}

We present a system for online draft co-training in RL post-training under context and pipeline parallelism. The system is built on two mechanisms: packed zigzag-ring attention, which supports branch-structured draft attention under CP, and TapChannel, which delivers target features across pipeline stages outside the pipeline schedule. Across models from 8B to 122B and both single- and multi-turn workloads, co-training keeps the policy's accuracy close to the no-draft baseline while accelerating end-to-end training by 1.19--1.88$\times$. Our CP attention is up to 2.9$\times$ faster than USP with a 2.7$\times$ reduction in per-GPU memory, and TapChannel's exposed wait stays within 5\% of update time. Future work includes improving speculation speedups on linear-attention targets, where they are currently smaller.

%% file: sec/appendix.tex

\clearpage
\section{Experimental Setup}
\label{app:exp-config}

\paragraph{Targets and drafts.}
We evaluate three draft families, EAGLE-3~\citep{li2026eagle}, DFlash~\citep{chen2026dflash}, and DSpark~\citep{cheng2026dspark}, all with Qwen3-8B~\citep{qwen3} as the target. To test scalability and generality, we also evaluate larger targets: Qwen3.5-27B, Qwen3.5-35B-A3B, Qwen3.5-122B-A10B~\citep{qwen35}, and GPT-OSS-120B with DFlash; Qwen3-14B and Nemotron-3.5-Lightning-30B-A3B~\citep{nvidia_nemotron_3_super} with DSpark. All drafts are initialized from official checkpoints. The inference engines serve the draft being co-trained: its weights are synchronized into the engines together with the policy weights at every refit. EAGLE-3, DFlash, and DSpark propose 3, 15, and 7 speculative tokens per verification step, respectively, except that DFlash proposes 9 on GPT-OSS-120B, and is trained with 15 but served with 7 on Qwen3.5-35B-A3B and Qwen3.5-122B-A10B.

\begin{table*}[h]
  \centering
  \caption{Training configuration of each run. TP/PP/CP/EP are the tensor-, pipeline-, context-, and expert-parallel degrees; responses per update are prompts $\times$ samples per prompt. All runs use GRPO with a vLLM engine colocated with training.}
  \label{tab:exp-config}
  \resizebox{\textwidth}{!}{%
  \begin{tabular}{@{}lllccc@{}}
    \toprule
    Target & Draft & Task & GPUs & TP/PP/CP/EP & Responses $\times$ Seq.\ len \\
    \midrule
    Qwen3-8B  & EAGLE-3, DFlash, DSpark & Math      & 32 H100 & 2/1/1/1 & 32$\times$16, 20K \\
    Qwen3-8B  & EAGLE-3, DFlash, DSpark & Workplace & 8 H100  & 2/1/2/1 & 32$\times$8, 32K \\
    Qwen3-14B & DSpark                  & Math      & 32 H100 & 4/1/1/1 & 32$\times$16, 20K \\
    \midrule
    Qwen3.5-27B                    & DFlash & Math & 32 GB200 & 4/2/2/1 & 32$\times$16, 20K \\
    Qwen3.5-35B-A3B                & DFlash & Math & 32 H100  & 2/2/2/8 & 32$\times$16, 20K \\
    Nemotron-3.5-Lightning-30B-A3B & DSpark & Math & 32 GB200 & 2/2/2/8 & 32$\times$16, 20K \\
    \midrule
    Qwen3.5-122B-A10B & DFlash & Math & 64 GB200 & 4/4/2/16 & 32$\times$8, 20K \\
    GPT-OSS-120B      & DFlash & Math & 64 GB200 & 4/4/2/16 & 32$\times$8, 20K \\
    \bottomrule
  \end{tabular}}%
\end{table*}

\paragraph{Tasks.}
For single-turn reasoning, we train on DAPOMath-17K~\citep{yu2025dapo} with 4k input and 16k output tokens and evaluate on AIME~2024~\citep{aime24}. Validation runs every 5 policy updates with 32 samples per problem (960 responses) for Qwen3-8B and Qwen3-14B, and every 50 updates with 16 samples per problem (480 responses) for the larger targets. For multi-turn evaluation, we adapt the NeMo Gym Workplace Assistant~\citep{styles2024workbench} on Qwen3-8B, with a sequence length of 32{,}768 tokens including tool feedback. Workplace Assistant is a multi-turn, agentic tool-use environment within NVIDIA NeMo Gym, designed for complex task execution in simulated office scenarios. It is based on the WorkBench benchmark~\citep{styles2024workbench}, a simulated workplace tool-use environment with email, calendar, CRM, project management, and analytics databases.

\paragraph{Training.}
All experiments run in NeMo-RL~\citep{nemo-rl} with GRPO~\citep{shao2024deepseekmath} and a vLLM~\citep{kwon2023efficient} inference engine colocated with training. Each policy update draws 512 responses (32 prompts $\times$ 16), or 256 (32 $\times$ 8) for Qwen3.5-122B-A10B, GPT-OSS-120B, and the Workplace Assistant, and takes one optimizer step over all of them. Rollouts are sampled at temperature 1.0 and top-$p$ 1.0. The policy is trained in BF16 with Adam at a constant learning rate of $10^{-6}$ after 5 warm-up updates, weight decay 0.1, and gradient clipping at 1.0. The policy loss is aggregated at the token level without a KL penalty. On the math task, it uses asymmetric clipping ($\epsilon_{\mathrm{low}}=0.2$, $\epsilon_{\mathrm{high}}=0.28$) and a truncated importance-sampling correction capped at 2; on the Workplace Assistant, it uses symmetric clipping at 0.2 without importance-sampling correction. The draft is optimized jointly with the policy ($\lambda=1$) at a learning rate of $3\times10^{-5}$ ($5\times10^{-5}$ for EAGLE-3) with weight decay 0.01. DFlash and DSpark sample 64 anchors per sequence. EAGLE-3 is trained with 3 TTT passes.

\paragraph{DSpark Heads under Context Parallelism.}
\label{app:dspark-cp}

DSpark adds a Markov head and a confidence head to the DFlash backbone~\citep{cheng2026dspark}. In training, slot $j$ of a $\gamma$-slot block anchored at $p$ predicts $x_{p+j+1}$, and both heads read the ground-truth token $x_{p+j}$. Under CP, the block runs on its anchor's rank, but its positions can extend into the next zigzag chunk. Every CP rank keeps the unsharded token IDs of the microbatch, so the heads look up their inputs and labels locally, without communication. The only cross-rank input is the target distribution for the TV loss, which extends at most $\gamma-1$ positions past the anchor's chunk and is received from the rank holding the next chunk, as in the DFlash loss.

\paragraph{Hardware and parallelism.}
On H100, Qwen3-8B trains with TP2, Qwen3-14B trains with TP4, and Qwen3.5-35B-A3B trains with TP/PP/CP/EP $=$ 2/2/2/8. On GB200, Qwen3.5-27B trains with TP/PP/CP $=$ 4/2/2, Nemotron-3.5-Lightning-30B-A3B trains with TP/PP/CP/EP $=$ 2/2/2/8, and Qwen3.5-122B-A10B and GPT-OSS-120B train with TP/PP/CP/EP $=$ 4/4/2/16. The Workplace Assistant runs with TP2 and CP2. Each vLLM engine uses TP1 for Qwen3-8B and Qwen3.5-27B, TP2 for Qwen3-14B, Qwen3.5-35B-A3B, and Nemotron-3.5-Lightning-30B-A3B, and TP4 for Qwen3.5-122B-A10B and GPT-OSS-120B. Table~\ref{tab:exp-config} summarizes the configuration.

\paragraph{Metrics.}
We report training reward, validation accuracy, and training--inference KL divergence: the per-token KL (K3 estimator) between the log-probabilities from the training and inference backends under identical weights, where low values indicate numerical consistency. For speculative decoding, we report acceptance length, the mean number of tokens accepted per verification. We also report rollout throughput, rollout speedup relative to the baseline, and end-to-end speedup in total policy-update time.

\section{Effect of Rollout Concurrency}
We evaluate Qwen3-8B with 1--128 concurrent sequences on H100, as shown in Figure~\ref{fig:rollout-concurrency}. Acceptance length remains nearly constant across concurrency, yet speculative decoding speedups decline as concurrency increases. Higher acceptance therefore does not guarantee greater rollout speedup; the serving configuration also determines how much benefit speculative decoding delivers. In particular, serving a fixed rollout batch with fewer engines can increase per-engine concurrency and reduce the resulting speedup. With 8{,}192-token responses, the speedups of EAGLE-3, DFlash, and DSpark are 1.69$\times$, 2.06$\times$, and 2.54$\times$ at 16 sequences per engine, and 1.72$\times$, 1.82$\times$, and 2.47$\times$ at 32.
\input{tex_figure/rollout_concurrency}

\section{Co-Training Overhead under Pipeline Parallelism}
\label{subsec:exp-pp}

Table~\ref{tab:pp-overhead} compares Qwen3-8B co-training with a no-draft baseline of the same parallel layout at PP$=2$ and PP$=4$. Rollout speedups remain similar across the two pipeline depths, while the added update time approximately doubles at PP$=4$. Increasing PP from 2 to 4 halves the number of policy replicas, so each draft stage processes twice as many microbatches per update. EAGLE-3 incurs more overhead than DFlash and DSpark because it requires multiple TTT passes. Waiting for target features accounts for 0.6--0.8\,s of this per update at PP$=2$ and 1.3--2.1\,s at PP$=4$ (Table~\ref{tab:tap-channel-runs}); the rest is the draft's own forward and backward on the last stage. Rollout savings outweigh the added training cost, yielding 1.28--1.85$\times$ end-to-end speedups across all configurations.
\input{tex_table/pp_overhead_table}

\section{Stage Times under Pipeline Parallelism}
\label{app:pp-balance}
\input{tex_table/pp_balance_table}

Table~\ref{tab:pp-balance} lists the per-stage times, periods and update times behind Figure~\ref{fig:pp-balance}, together with two layouts of Qwen3.5-122B-A10B. 

\section{Correctness Checks}
\label{app:correctness}

Our context-parallel attention is covered by unit tests that compare the distributed path against a single-rank reference on the same inputs. For block-draft attention, the packed zigzag ring at CP$=2$ and CP$=4$ reproduces the single-rank output to $2\times10^{-2}$ and its gradients to $5\times10^{-2}$ in bf16, including the sliding-window variant; the DSpark head reproduces logits and confidences to $3\times10^{-2}$ and its loss to $2\times10^{-3}$. For EAGLE-3 TTT, the two-part flash decomposition with its log-sum-exp merge matches a dense reference to $3\times10^{-2}$, and the packed path reproduces the unpacked loss and gradients to $10^{-5}$. The USP baseline of \S\ref{subsec:exp-cp} is checked the same way, against a single-GPU reference, for every Ulysses/ring factorization it is run with. These are small-shape tests, tens to hundreds of tokens per sequence, so they verify the algebra of the sharding and of the merge, not numerical behavior at 128K.

\section{Per-Phase Step Times}
\label{app:step-cost}

Table~\ref{tab:step-cost} breaks down the time of a policy update for all six workloads with complete step timers. Rollout dominates the step without a draft, taking 66--73\% of the time on the math task and 56\% on the Workplace Assistant, where rollout also includes tool execution. Speculative decoding shortens rollout by 1.27--2.39$\times$ on the math task and by 1.79--2.37$\times$ on the Workplace Assistant. Co-training lengthens the update by 6--12\% on the math task and by 14--20\% on the Workplace Assistant. Refit grows by at most 7\%. The other column shrinks by 28--35\,s on Qwen3-8B and Qwen3-14B with the math task, whose validation runs every 5 updates and is also generated with speculative decoding; on the other workloads, it changes by at most 1.2\,s. The rollout savings far exceed the added update time, giving end-to-end speedups of 1.13--2.02$\times$ on the math task. On the Workplace Assistant, the speedups are 1.26--1.41$\times$, smaller than its rollout speedups, because rollout takes only 56\% of the step there and co-training lengthens the update the most.

\input{tex_table/step_cost_table}

\section{Training Curves of Additional Targets}
\label{app:curves}
\input{tex_figure/appendix_curves}

Figure~\ref{fig:appendix-curves} shows the training curves of the six targets in Table~\ref{tab:main-performance} other than Qwen3-8B, each against a baseline trained without a draft. The co-trained runs follow the baseline reward on all six targets. Training--inference KL divergence stays on the same order of magnitude as the baseline ($10^{-4}$--$10^{-3}$). Rollout throughput is higher than the baseline throughout training. End-to-end step time is lower for the co-trained run on most updates.

On the MoE and linear-attention targets, rollout speedups are smaller than on the dense Qwen3-8B and Qwen3-14B, although acceptance lengths are similar or longer (Table~\ref{tab:main-performance}). Verifying several tokens is relatively more expensive on these architectures: on MoE targets the verified tokens activate more experts~\citep{huang2026moesd}, and on linear-attention targets autoregressive decoding is already cheap, leaving less for speculation to save~\citep{wang2026specla}. Reducing the verification cost on these architectures is a direction for future work.

%% file: tex_figure/rollout_concurrency.tex
\begin{figure}[t]
  \centering
  \includegraphics[width=\linewidth]{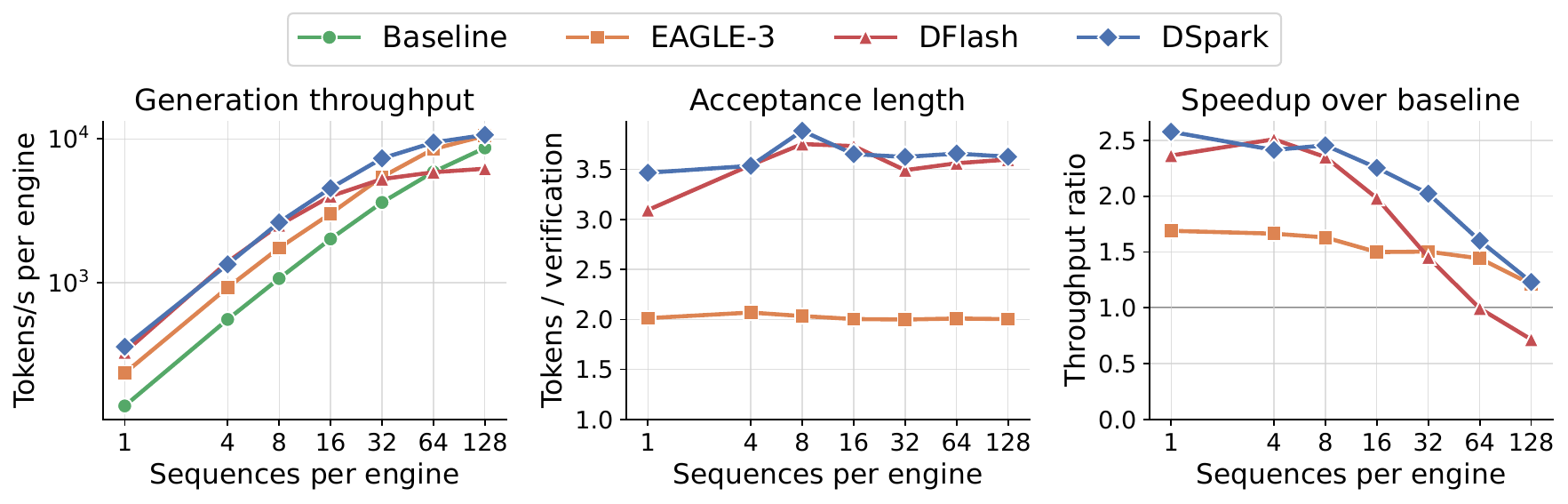}
  \vspace{-2em}
  \caption{Effect of per-engine concurrency on Qwen3-8B generation. Left: generation throughput, counting only target tokens. Middle: acceptance length. Right: throughput relative to AR decoding. Speculative decoding speedups decline as concurrency increases despite nearly constant acceptance length.}
  \label{fig:rollout-concurrency}
\end{figure}

%% file: tex_table/pp_overhead_table.tex
\begin{table}[t]
  \centering
  \caption{Pipeline-parallel overhead for Qwen3-8B, averaged over first 10 policy update steps. DSpark achieves the largest end-to-end speedup (1.85$\times$) with 14.6\% update-time overhead.}
  \label{tab:pp-overhead}
  \scriptsize
  \begin{tabular}{lcccccc}
    \toprule
    & & \multicolumn{2}{c}{Rollout} & \multicolumn{2}{c}{Train} & \\
    \cmidrule(lr){3-4} \cmidrule(lr){5-6}
    Draft & AL $\uparrow$ & Time (s) & Speedup $\uparrow$ &
    Time (s) & Overhead $\downarrow$ & E2E $\uparrow$ \\
    \midrule
    \multicolumn{7}{l}{\emph{PP$=2$}} \\
    None & - & 329.2 & 1.00$\times$ & 31.5 & - & 1.00$\times$ \\
    EAGLE-3 & 1.89 & 238.7 & 1.38$\times$ & 42.4 & 34.3\% & 1.31$\times$ \\
    DFlash & 2.84 & 229.2 & 1.44$\times$ & 35.8 & 13.6\% & 1.37$\times$ \\
    DSpark & 3.37 & 156.8 & 2.10$\times$ & 36.1 & 14.6\% & 1.85$\times$ \\
    \midrule
    \multicolumn{7}{l}{\emph{PP$=4$}} \\
    None & - & 328.3 & 1.00$\times$ & 34.8 & - & 1.00$\times$ \\
    EAGLE-3 & 1.89 & 232.7 & 1.41$\times$ & 53.9 & 54.9\% & 1.28$\times$ \\
    DFlash & 2.84 & 229.4 & 1.43$\times$ & 43.6 & 25.3\% & 1.34$\times$ \\
    DSpark & 3.36 & 157.1 & 2.09$\times$ & 43.5 & 25.1\% & 1.79$\times$ \\
    \bottomrule
  \end{tabular}
\end{table}

%% file: tex_table/pp_balance_table.tex
\begin{table*}[t]
  \centering
  \caption{Stage balance under co-training at PP$=4$. Stage time is forward-backward GPU time per microbatch; period is wall time per microbatch during steady-state 1F1B; update is wall time per policy update. Rel.\ reports update time relative to the even-layout baseline: no-draft training for Qwen3.5-35B-A3B and co-training for Qwen3.5-122B-A10B.}
  \label{tab:pp-balance}
  \small
  \setlength{\tabcolsep}{9pt}
  \begin{tabular}{@{}llccccccc@{}}
    \toprule
    \multirow{2}{*}{Layers per stage}
    & \multirow{2}{*}{Draft}
    & \multicolumn{4}{c}{Stage time (ms)}
    & \multirow{2}{*}{Period (ms) $\downarrow$}
    & \multirow{2}{*}{Update (s) $\downarrow$}
    & \multirow{2}{*}{Rel.\ $\downarrow$} \\
    \cmidrule(lr){3-6}
    & & 0 & 1 & 2 & 3 (draft) & & & \\
    \midrule
    \multicolumn{9}{l}{\emph{Qwen3.5-35B-A3B, 40 layers}} \\
    10/10/10/10 & None & 486 & 470 & 445 & 620 & 630 & 76.3 & 1.00$\times$ \\
    10/10/10/10 & DFlash    & 558 & 494 & 463 & 740 & 757 & 89.5 & 1.17$\times$ \\
    11/11/11/7  & DFlash    & 558 & 517 & 495 & 594 & 622 & 73.3 & 0.96$\times$ \\
    11/11/11/7  & None & 528 & 505 & 493 & 474 & 548 & 65.2 & 0.85$\times$ \\
    \midrule
    \multicolumn{9}{l}{\emph{Qwen3.5-122B-A10B, 48 layers}} \\
    12/12/12/12 & DFlash    & 507 & 486 & 506 & 632 & 625 & 74.4 & 1.00$\times$ \\
    13/13/13/9  & DFlash    & 504 & 522 & 529 & 491 & 560 & 67.6 & 0.91$\times$ \\
    \bottomrule
  \end{tabular}
\end{table*}

%% file: tex_table/step_cost_table.tex
\begin{table*}[t]
  \caption{Time per policy update by phase (s), without and with draft co-training.  Other includes validation, log-probability recomputation, and data and reward processing.  E2E is the ratio of total step time against the run without a draft.}
  \label{tab:step-cost}
  \centering
  \small
  \setlength{\tabcolsep}{4pt}   
  \begin{tabular*}{\textwidth}{@{\extracolsep{\fill}}lcrrrrrc@{}}
    \toprule
    Draft & AL $\uparrow$ & Rollout & Update & Refit & Other & Total $\downarrow$ & E2E $\uparrow$ \\
    \midrule
    \multicolumn{8}{l}{\emph{Qwen3-8B, Math}} \\
    None & --- & 279.5 & 23.7 & 6.8 & 75.3 & 385.3 & 1.00$\times$ \\
    EAGLE-3 & 2.27 & 163.6 & 26.6 & 7.0 & 47.6 & 244.8 & 1.57$\times$ \\
    DFlash & 3.43 & 123.6 & 25.0 & 6.9 & 41.9 & 197.6 & 1.95$\times$ \\
    DSpark & 3.66 & 117.1 & 26.3 & 6.9 & 40.1 & 190.5 & 2.02$\times$ \\
    \midrule
    \multicolumn{8}{l}{\emph{Qwen3-8B, Workplace Assistant}} \\
    None & --- & 78.7 & 35.4 & 9.2 & 16.1 & 139.4 & 1.00$\times$ \\
    EAGLE-3 & 2.46 & 43.9 & 42.6 & 9.4 & 15.1 & 111.0 & 1.26$\times$ \\
    DFlash & 2.85 & 42.8 & 41.5 & 9.8 & 16.1 & 110.2 & 1.26$\times$ \\
    DSpark & 3.50 & 33.2 & 40.4 & 9.5 & 15.4 & 98.6 & 1.41$\times$ \\
    \midrule
    \multicolumn{8}{l}{\emph{Qwen3-14B, Math}} \\
    None & --- & 245.8 & 43.2 & 7.5 & 77.0 & 373.5 & 1.00$\times$ \\
    DSpark & 3.62 & 107.9 & 46.5 & 7.5 & 46.9 & 208.9 & 1.79$\times$ \\
    \midrule
    \multicolumn{8}{l}{\emph{Qwen3.5-27B, Math}} \\
    None & --- & 245.8 & 65.6 & 5.0 & 21.3 & 337.6 & 1.00$\times$ \\
    DFlash & 5.10 & 167.5 & 72.1 & 5.2 & 22.0 & 266.8 & 1.27$\times$ \\
    \midrule
    \multicolumn{8}{l}{\emph{Qwen3.5-35B-A3B, Math}} \\
    None & --- & 133.5 & 39.9 & 6.8 & 16.5 & 196.7 & 1.00$\times$ \\
    DFlash & 4.24 & 105.1 & 44.8 & 7.0 & 16.6 & 173.6 & 1.13$\times$ \\
    \midrule
    \multicolumn{8}{l}{\emph{Qwen3.5-122B-A10B, Math}} \\
    None & --- & 105.9 & 31.6 & 7.5 & 15.5 & 160.5 & 1.00$\times$ \\
    DFlash & 4.72 & 62.3 & 31.3 & 7.5 & 14.3 & 115.5 & 1.39$\times$ \\
    \bottomrule
  \end{tabular*}
\end{table*}

%% file: tex_figure/appendix_curves.tex
\begin{figure*}[!t]
  \centering
  \includegraphics[width=\linewidth]{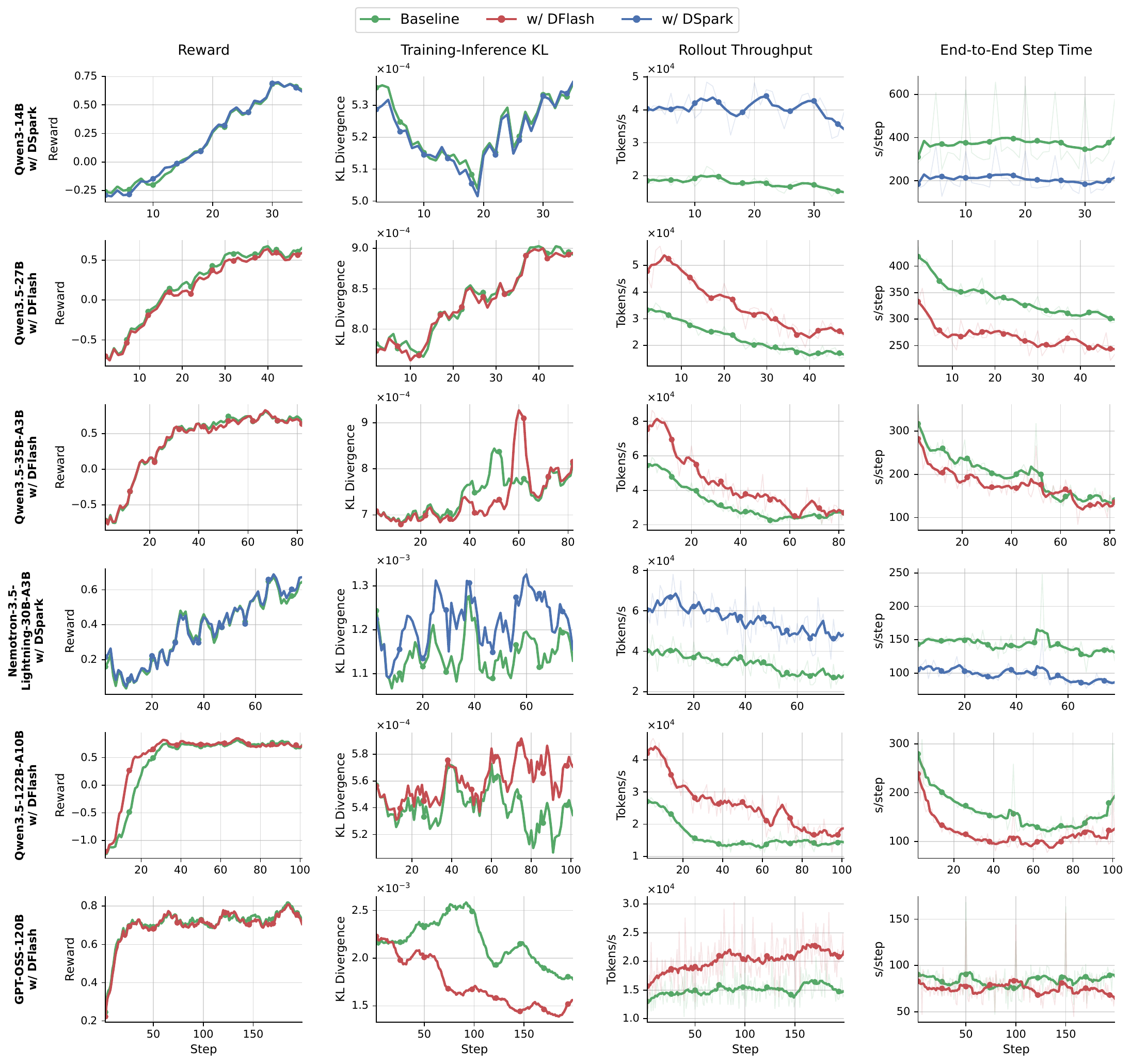}
  \caption{Training curves of the remaining targets in Table~\ref{tab:main-performance}.
  Each row compares a target trained with and without draft co-training. Curves are centered moving averages;
  translucent lines show raw rollout throughput and end-to-end step time.}
  \label{fig:appendix-curves}
\end{figure*}